\documentclass[conference,letter]{IEEEtran}

\usepackage{cite}
\usepackage{amsmath,amssymb,amsfonts}
\usepackage{algorithmic}
\usepackage{textcomp}
\usepackage{xcolor}
\def\BibTeX{{\rm B\kern-.05em{\sc i\kern-.025em b}\kern-.08em
    T\kern-.1667em\lower.7ex\hbox{E}\kern-.125emX}}
\usepackage{placeins}
\usepackage{graphicx}
\usepackage{subcaption}

\usepackage{float}

\usepackage{amsmath,amssymb,amsfonts}

\usepackage{hyperref}
\usepackage{url}

\usepackage{xurl}

\usepackage{tikz}
\usetikzlibrary{positioning}
\newdimen\nodeDist
\usepackage{tikzlings}
\usepackage{lmodern}
\usepackage{tikzpeople}
\usetikzlibrary{positioning,shadows.blur,shapes.geometric,calc,fit}
\tikzset{database/.style={cylinder,aspect=1,draw,fill,shape border rotate=90,path picture={
\draw[white] let \p1=($(path picture bounding box.north east)-(path picture bounding
box.south west)$) in 
foreach \XX in {1,2,3}  {([yshift=-\XX*\y1/4]path picture bounding box.north west) 
arc(180:360:\x1/2 and 0.25*\x1/2)};
}}}

\usetikzlibrary{positioning}

\newtheorem{example}{Example}
\newtheorem{remark}{Remark}
  
\begin{document}

\date{}





 \title{One Picture is Worth a Thousand Words: \\A New Wallet Recovery Process
 }


\author{\IEEEauthorblockN{Hervé Chabanne}
\IEEEauthorblockA{\textit{IDEMIA and Télécom Paris} \\
herve.chabanne@telecom-paris.fr}
\and
\IEEEauthorblockN{Vincent Despiegel}
\IEEEauthorblockA{\textit{IDEMIA}}
\and
\IEEEauthorblockN{Linda Guiga}
\IEEEauthorblockA{\textit{IDEMIA and Télécom Paris}}
}

\maketitle

\begin{abstract}
We introduce a new wallet recovery process. Our solution associates 1) visual passwords: a photograph of a secretly picked object (Chabanne et al., 2013) with  2) ImageNet classifiers transforming images into binary vectors and, 3) obfuscated fuzzy matching (Galbraith and Zobernig, 2019) for the storage of visual passwords/retrieval of wallet seeds. Our experiments show that the replacement of long seed phrases by a  photograph is possible.
\end{abstract} 


\section{Introduction}\label{introduction}
Cryptocurrency wallets store  private keys and make use of them for performing  transactions among blockchains. Their loss  is identified in \cite{DBLP:journals/comsur/ContiELR18} as one of the three challenges associated to Bitcoin. Today, they mainly rely on a seed phrase for their recovery \cite{bipMnemonic}. In 2021, the New York Times reported \cite{NYT} that  ``20 percent of the existing 18.5 million Bitcoin or around 3.7 million BTCs appear to be lost due to forgotten passwords''. 

To alleviate the burden of remembering this long password, we alternatively rely on the concept of visual passwords introduced in 2013 by Chabanne et al. \cite{DBLP:journals/iacr/ChabanneCDFN13}. 

The underlying principle of visual passwords is, in the context of authentication, the following:
\begin{itemize}
\item At the registration step, you choose an object and take a photograph of it. Your  choice has to remain secret.
\item When you want to authenticate yourself, you take another photograph of the same object for a comparison image vs image with the reference.
\end{itemize}

While \cite{DBLP:journals/iacr/ChabanneCDFN13} focuses on a single type of object: Hamiltonian circuits among a cube, with a design enabling many possible configurations;  to ensure a good entropy, we here let the users  choose among a great variety of different objects. 

A special care is  taken to the storage of references. We apply the work of Galbraith and Zobernig \cite{DBLP:conf/tcc/GalbraithZ19} to perform 
Hamming ball membership  determining  in an obfuscated way whether a binary vector lies close to a predetermined center.

Our main contribution is the introduction of a novel wallet recovery system. Moreover, we show its feasibility by our experiments transforming visual passwords by  state of the art image processing algorithms into suitable binary vectors, called \emph{templates} in the following, for secure storage/ seed retrieval.

The rest of the paper is organized as follows: in Sec. \ref{obfucation} we recall the techniques and security properties of  obfuscated Hamming distance comparisons and show how to deliver a payload in case of a matching. In Sec. \ref{vgg}, we describe how to transform  visual password pictures into binary vector templates thanks to deep learning algorithms. In Sec. \ref{experiments}, we report our experiments. Sec. \ref{proposal} 
details our proposal. Sec. \ref{conclusion} concludes.

\subsection{Related Works}
For a general introduction to wallets in the context of Bitcoin; see, for instance,  chapter 4 of \cite{books/daglib/0040621}.

While there are numerous other attempts to replace passwords using graphical interfaces and images \cite{DBLP:journals/csur/BiddleCO12}, visual passwords \cite{DBLP:journals/iacr/ChabanneCDFN13} share  a lot with biometric recognition. For instance, we are using  in Sec. \ref{experiments} the same tools to evaluate the accuracy of our proposal, namely:
\begin{itemize}
    \item  The \emph{False Acceptance Rate} (FAR) measures  the proportion of times an imposter can fool the system. FAR is directly related to the security level.
    \item On the opposite, the user's convenience is gauged thanks to the \emph{False Reject Rate} (FRR)  which corresponds to genuine attempts dismissed.
\end{itemize}
As one cannot win  both at the same time with FAR and FRR, the  \emph{Detection Error Tradeoff (DET)} curve which represents  false rejection rate vs. false acceptance rate comes into consideration for determining the \emph{Equal Error Rate (EER)} of the system where FAR and FRR are equal. 

Major differences  however differentiate biometrics and visual passwords. Biometrics are public and immutably linked to a person while  visual passwords are secret and easy to renew. For instance, biometrics need liveness anti-spoofing countermeasures to thwart impersonation attacks and depending on their application, privacy enhancing technologies are necessary too.

Fuzzy matching has already been considered in the context of biometrics, around the notion of \emph{secure sketch} introduced in \cite{DBLP:conf/eurocrypt/DodisRS04}. Here the matching is realized thanks to an underlying error correcting code. As indicated in \cite{DBLP:conf/tcc/GalbraithZ19}, parameters of secure sketch are thus ``strongly constrained by the
need for an efficient decoding algorithm''. Moreover, when it comes to their implementation with real (biometric) data, their security is questionable \cite{DBLP:conf/sp/SimoensTP09}, in particular regarding their reusability.

Cryptographic techniques such as a secret sharing mechanism \cite{DBLP:conf/eurosp/JareckiKKX16} or multi-signature techniques \cite{DBLP:conf/sac/HanSES21} can also be envisaged for cryptocurrency wallets. 
Our proposal is -- the same way, that passphrases are -- complementary.
Another solution \cite{DBLP:conf/globecom/RezaeighalehZ19} relies on a Diffie-Hellman exchange between hardware wallets with a human visual verification to thwart man-in-the-middle attacks.

\section{Obfuscated Fuzzy Hamming Distance Matching}\label{obfucation}
\begin{figure}\centering
\includegraphics[width=\columnwidth]{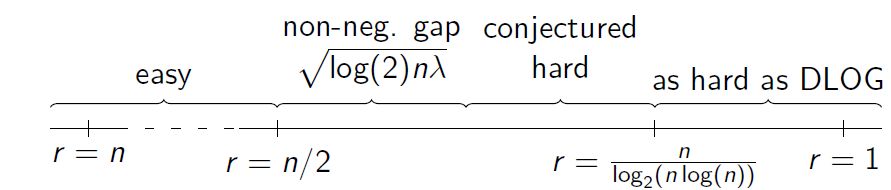}
\caption{Hardness of MSP Problem 
}
\label{hard}
\end{figure}
In this section, we show how to store our reference binary vector templates in a way that enables Hamming distance comparisons while preserving their confidentiality. I.e. we retrieve the wallet's seed when and only when a fresh template close to the reference is entered.
For that, we make use of cryptographic obfuscation.

Obfuscation makes programs  unintelligible  while preserving their functionality.
General obfuscation techniques are either impossible  \cite{DBLP:conf/crypto/BarakGIRSVY01} or, despite major progress \cite{DBLP:conf/stoc/JainLS21}, ineffective.
In 2014, \cite{DBLP:conf/tcc/BarakBCKPS14} defines practical input-hiding obfuscation techniques for evasive functions including \emph{point functions} ``$x==e$'', which return 1 when the input is equal to a predetermined constant $e$ and 0 otherwise.

\begin{example}\label{ex1}
\cite{DBLP:conf/trustcom/ZobernigGR19} describes how to obfuscate variable comparisons ``$ax+b==y$'' where $a,b$ are two $k$-bits constants. Let $H$ stand for a preimage-resistant hash function with $n$-bits outputs, $n > k$. Choose at random $t \in \{0,1\}^{n-k}$ and $r \in \{0,1\}^k$.  Let $h=H(r||t) $ and $u=r+b$. Values $a,u,h$ are published. The obfuscated program then checks \begin{equation} \label{eq1} H(ax + u - y||t)==h\end{equation} while keeping the value $b$ hidden. Note that when (\ref{eq1}) is verified by inputs $(x,y)$, $b$ can be retrieved as $b=y-ax$.
\end{example}

Relying on a number-theoretic computational assumption  called the Modular Subset Product (MSP) problem (see Fig. \ref{hard}), \cite{DBLP:conf/tcc/GalbraithZ19}  defines a Hamming distance obfuscator which checks whether an $n$-bits binary vector $x$ is within Hamming distance $r$ of a predetermined $c$ for 
\begin{equation}\label{r1}
    r\leq n/2 - \sqrt{\log(2) n \lambda}  
\end{equation}
where $\lambda$ is a security parameter.

A  vector $c = (c_1,\ldots, c_n) \in \{0,1\}^n$ is encoded as \begin{equation}\label{encode}
\text{ENCODE}(c)=((p_i)_{i=1,\ldots,n},q,C)
\end{equation} where
\begin{itemize}
    \item 
$
    C = \prod_{i=1}^{n}p_i^{c_i} \text{mod}~ q $;
    \item $(p_i)_{i=1,\ldots,n}$ are small distinct primes taken at random for each encoding;
    \item $q$ is a small safe prime verifing  $\prod_{i \in I} < q/2$ for all $I \subset {1,\ldots,n}$ with cardinality $|I| < r$. Typically, $q \sim (n\log n)^r$.
    \end{itemize}
This encoding procedure keeps the vector $c$ hidden when \begin{equation}\label{r2}
r>log(2\sqrt{2\pi e})\dfrac{n}{log(nlog(n))}
\end{equation}

A procedure DECODE is then defined s.t. DECODE$((p_i)_{i=1,\ldots,n},q,C,x)=c$ for each vector $x$ which stands at Hamming distance $d(c,x)<r$. This procedure returns $\perp$ for  $x$ s.t. $d(c,x)\geq r$ except when a false acceptance occurs. Note that, when $r$ satisfies (\ref{r2}), this false acceptance cannot happen,
see Sec. \ref{Implemen} and \cite{DBLP:conf/tcc/GalbraithZ19} for details.

The obfuscated program with embedded data $(p_i)_{i=1,\ldots,n},q,C$ is executed as follows for an input $x$:

\noindent 
\texttt{l$_1$:}
$c'=\text{DECODE}((p_i)_{i=1,\ldots,n},q,C,x)$

\noindent 
\texttt{l$_2$:}
If $c'=\perp$ Return $0$

\noindent 
\texttt{l$_3$:}
Return the obfuscated point function comparison to $c$

In \cite{DBLP:conf/tcc/GalbraithZ19}, the last line 
\texttt{l$_3$}
eliminates false acceptances. 
Write now $c=c_1||c_2$ and for $b$ at random, $a c_1+b=c_2$.  In our proposal, we replace
\texttt{l$_3$}
by (using the notations introduced in Example \ref{ex1}):

\noindent \texttt{l$_3$':} If (\ref{eq1}) stands for inputs $c_1'||c_2'=c'$ Return $b=c_2'-ac_1'$; else Return $\perp$

We then obtain the RETRIEVESEED program comprised of the three lines: \texttt{l$_1$}, \texttt{l$_2$}, \texttt{l$_3$'}.

\section{Pictures Processing}\label{vgg}
In this section, we describe our choices for transforming photographs of visual passwords into binary vector templates. Our experiments are reported in the next section.
\subsection{Templates Construction}

\begin{figure}\centering
\includegraphics[width=\columnwidth]{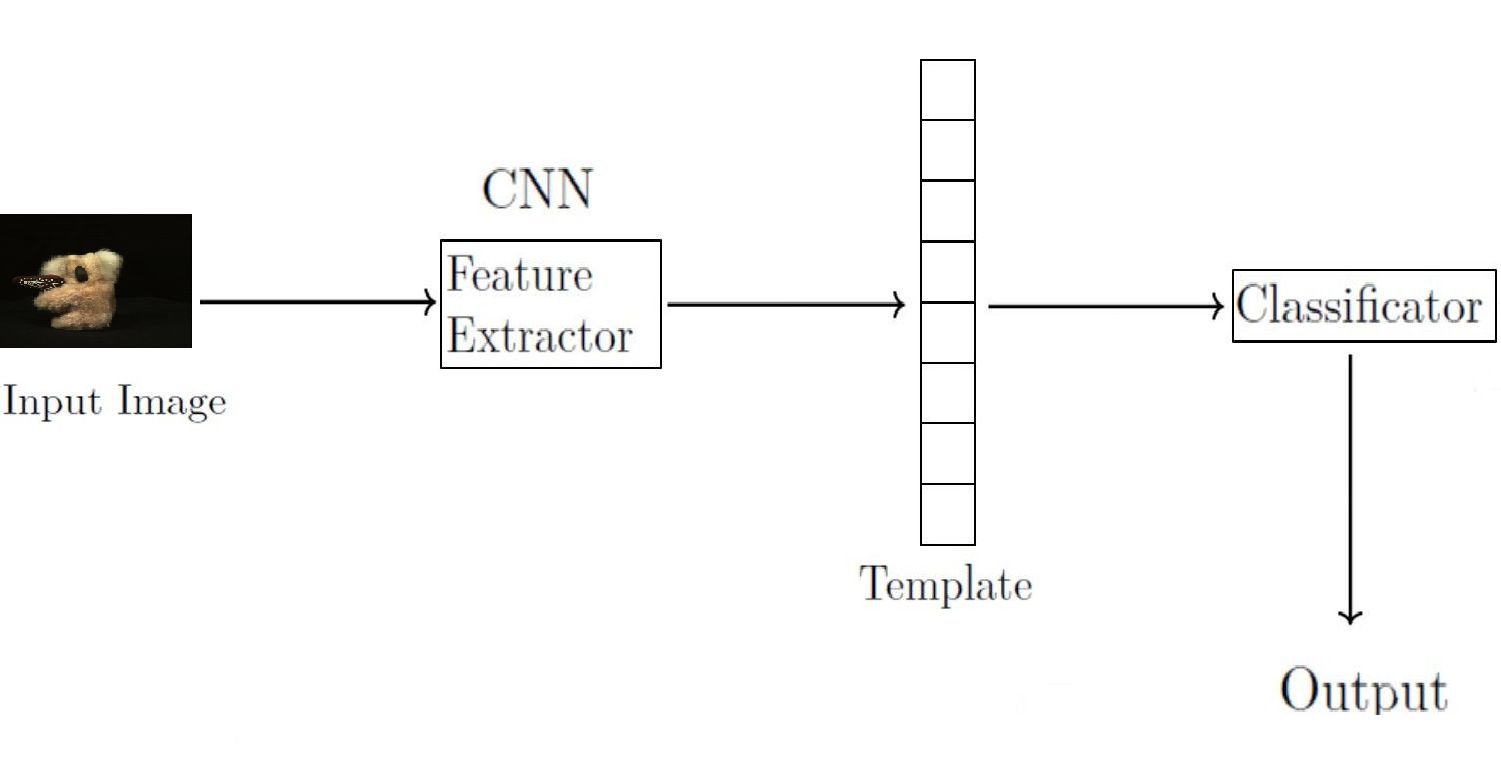}
\caption{An ImageNet Classifer}
\label{template}
\end{figure}

Consider the architecture of an ImageNet classifier as in Fig. \ref{template}. It takes as an input an image from which its features are extracted thanks to a Convolutional Neural Network (CNN) to eventually output a classification. Similarly to the idea used in Face Recognition algorithms, the underlying representation is a good candidate feature for object recognition even if the objects were not in the training dataset. Consequently, in a first step, we remove the last classification layers to just keep floating point vectors of the internal representation. Finally, we binarize these vectors to obtain our templates. 

\subsection{Model Choice}
We choose a model trained (with its parameters) among  \url{https://paperswithcode.com/sota/image-classification-on-imagenet}. 

After different trials (see Annex~\ref{selection}), we pick VGG-16 \cite{vgg} as the underlying model to classify images. This leads to  vectors with 4096 floating-point coordinates. To reduce the dimension to only 512 bits, we apply Locality Sensitive Hashing (LSH) \cite{DBLP:conf/stoc/IndykM98}. For this, we generate a random sparse matrix of shape $(4096, 512)$ thanks to the Scikit library: \url{https://scikit-learn.org/stable/}. We then multiply the generated matrix by the original coordinates. Lastly, we only keep the signs of the resulting vector elements so as to turn the floating values into binary ones.
At the end, our overall architecture is similar to the perceptual hashing algorithm NeuralHash \cite{NH}.

\section{Experiments}\label{experiments}

\subsection{Test Dataset}\label{aloi} 
\begin{figure}\centering
\includegraphics[width=\columnwidth]{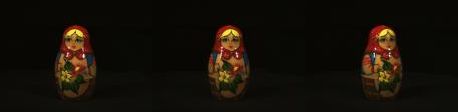}
\caption{Different Views of Object 197}
\label{souris}
\end{figure}

To validate our experiments, we use the Amsterdam Library of Object Images (ALOI) 
\cite{DBLP:journals/ijcv/GeusebroekBS05}. The ALOI dataset is made of 1,000 objects recorded under various viewing angles, illumination
angles, and illumination colors (Fig. \ref{souris}), yielding a total of 110,250 images  for the collection.

\subsection{Accuracy} \label{accuracy}
\begin{figure*}
\centering
\begin{subfigure}{.5\textwidth}
  \centering
\includegraphics[width=0.85\columnwidth]{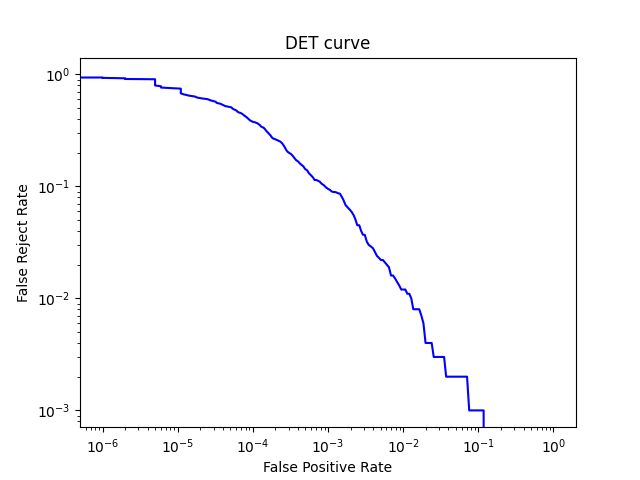} 
  \caption{Rotation of  35 degrees, EER=11$\%$}
  \label{fig:sub1}
\end{subfigure}%
\begin{subfigure}{.5\textwidth}
  \centering
\includegraphics[width=0.85\columnwidth]{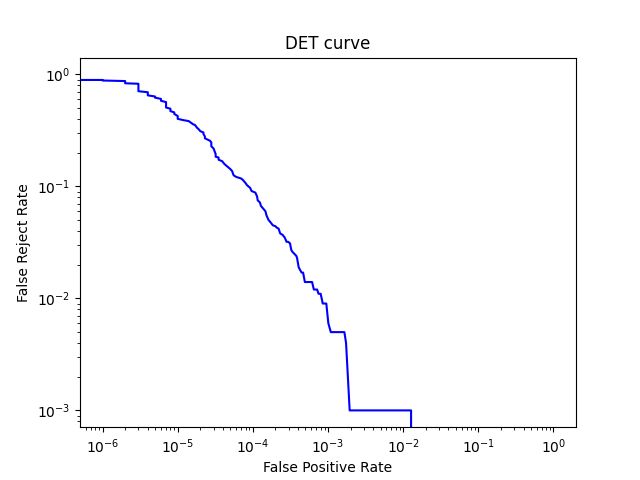} 
  \caption{Rotation of  15 degrees, EER=2$\%$}
  \label{fig:sub2}
\end{subfigure}
\caption{DET Curves}
\label{det}
\end{figure*}

To test the accuracy of our system, we select, for each object, 3 different views --  corresponding  to rotation angles of 0, 15 and 35 degrees -- which seems realistic in terms of noise for the target scenario. We then obtain the resulting DET curves shown in Fig. \ref{det}.

From our observations, our binarization process only slightly degrades our overall accuracy. Some more sophisticated methods as in \cite{DBLP:conf/asiaccs/UzunYCKL21} could be used for binarization but as the degradation is under control, our experiments stick to this simple method.

\subsection{Implementation Details} \label{Implemen}
With $n=512$, we choose $r=140$,  placing ourselves at the rightmost part of Fig. \ref{hard}. These parameters
satisfy both inequalities (\ref{r1}) and (\ref{r2}).

Our implementation of the ENCODE (resp. RETRIEVESEED) procedure  yields on our laptop  an average encoding time (resp. decoding time) of 50 ms (resp. 10 ms). These timings are in line with the ones given by \cite{DBLP:conf/tcc/GalbraithZ19}.

We obtain then an FAR around $4.10^{-4}$  for a rotation angle of 15 degrees (resp. 35 degrees) and a corresponding FRR of $1.8\%$ (resp. $7\%$). 
Note that, in our system -- in opposition to biometric systems where a false reject can imply for a user to be blocked at a gate -- a false reject
simply demands for a new photograph of the referenced object to be taken.

\begin{remark} A user can store different objects. For each of them, its encoding enables us to hide a new secret $k_i$, $i=1,\ldots,m$. By taking the exclusive OR of all of them $k_1 \oplus \ldots \oplus k_m=k$, the resulting $k$ is obtained for an illegitimate user when and only when 
each of the $m$ objects he had chosen leads to an encoding which matches with the genuine stored object.
\end{remark}

\section{Our Proposal}\label{proposal}

As pointed out by \cite{DBLP:conf/nips/AnsuiniLMZ19}, there is a huge gap, e.g. for VGG-16, regarding the intrinsic dimension -- i.e. the minimal number of parameters needed to describe a representation -- between input images and outputs of CNNs. We rely on that observation to 
mitigate the risk of an attack which looks for false positives.

We thus envisage to implement a proprietary algorithm on a dedicated server -- i.e. with dedicated metaparameters and training -- to compute templates corresponding to photos of objects. This way, template requests can be recorded and a security policy be established to limit their number, enforcing access control to templates.

To protect users against the server, we suggest sending visual passwords encrypted thanks to homomorphic encryption. Template computation is now possible directly in the encrypted domain \cite{DBLP:conf/cscml/ChillottiJP21}  and new progress is announced \cite{ro,DBLP:conf/micro/SamardzicFKDDP021}. 

We  now consider that a homomorphic encryption scheme is chosen \cite{HomomorphicEncryptionSecurityStandard} and that users  generate their own private key for this scheme. Note that they do not have to keep them.

\subsection{Detailed Description}\label{details}
Our system is made of:
\begin{itemize}
    \item users;
    \item a dedicated server $S$ in charge of computing templates for users from their visual password. 
\end{itemize}

The different steps of our wallet recovery process are summarized in Fig. \ref{proposition} 

\begin{figure*}\centering
\fbox{%
\parbox{0.975\linewidth}{
\begin{description}
\item[Setup:] ~\\
User $U$ picks his visual password ~\\
$U$ generates his  key  for the underlying  homomorphic encryption scheme and encrypts his visual  password ~\\
$U$ asks the dedicated server $S$ for the template associated to his visual password ~\\
$S$ computes $U$'s template, $c$,  encrypted ~\\
$U$  decrypts $c$, computes ENCODE($c$) (\ref{encode}) and stores it at a place of his choice
\item[Seed Recovery:] ~\\
$U$ gets back ENCODE($c$) and takes a new photography of his visual password, (optionally, if needed, $U$ generates a new homomorphic encryption key), encrypts it and sends the ciphertext to $S$ ~\\
$U$ asks $S$ for the template $x$ corresponding to this new photography~\\
$S$ computes $x$ in an encrypted form ~\\
$U$ decrypts $x$ and retrieves his wallet's seed as RETRIEVESEED(ENCODE($c$),$x$)
\end{description}
}
}
\caption{Our Wallet Recovery Process} \label{proposition}
\end{figure*}

More specifically, a user $U$ is going to ENCODE his template (\ref{encode}) and perform RETRIEVESEED using his own device; e.g. his mobile phone. This device also enables him  to capture and then, encrypt his visual password (resp. decrypt his template)  before sending it to $S$ (resp. after receiving it from $S$). We consider that all these operations performed on the device of $U$ are safe from attacks. 

The way his wallet's seed is used after its retrieval is out-of-scope of this paper.


Server $S$ is considered honest-but-curious regarding the requests from the users. It protects the implementation of the proprietary model $M$ and restricts templates computation. 

\subsection{Security Discussion}\label{secdiscut}
We consider three factors to be taken into account for the security of our wallet seeds recovery process:
\begin{itemize}
    \item Storage of the obfuscated template;
    \item Access to a templates construction algorithm;
    \item Choice diversity for visual passwords.
\end{itemize}

Our  threat model is simple: we want to protect against an adversary who tries to retrieve the wallet's seed by presenting a binary vector close to the stored reference template.



Having access to the know-how for transforming visual passwords into templates enables this adversary to attempt to obtain a false acceptance. Otherwise, we rely on the security provided by the obfuscated fuzzy Hamming distance matching. 
Searching in $\{0,1\}^{n}$, $n=512$  leads  to a  probability of $1/{2^\lambda}$ with $\lambda=87$ for $r=140$ (see (\ref{r1})) to find by chance a vector in the targeted Hamming ball.
In contrast, the knowledge of the underlying templates subspace 
enables the adversary to drastically reduce his efforts with a FAR of $4.10^{-4}$ (see also Annex \ref{annexe}). The confidentiality of the model $M$ 
is paramount for our wallet recovery process.

Regarding that point, besides server compromise, given oracle access to a neural network as in our proposal, model extraction attacks can be launched \cite{DBLP:conf/uss/TramerZJRR16}. Today, the best attacks 
\cite{DBLP:conf/sp/WangG18,DBLP:series/lncs/OhSF19,DBLP:conf/cvpr/OrekondySF19,DBLP:conf/crypto/CarliniJM20,DBLP:conf/uss/ChandrasekaranC20} against ImageNet classifiers seem unpractical. Note also that a first defense strategy keeping the model’s accuracy has been introduced in  \cite{DBLP:journals/corr/abs-2201-09243}.

\section{Conclusion}\label{conclusion}
\begin{figure}\centering
\includegraphics[width=\columnwidth]{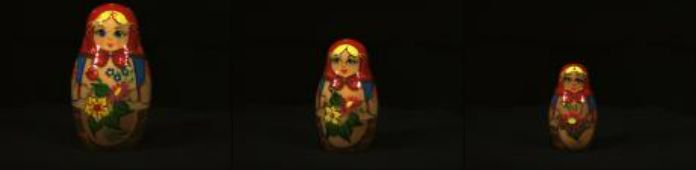}
\caption{Objects 196, 197, 198 side-by-side}
\label{dolls}
\end{figure}

We are confident that there is room for improvement of the accuracy of our system. For instance,  facial recognition which is an image processing problem of roughly the same difficulty as ours obtains less than $1\%$ of FRR at  FAR of $10^{-6}$ with systems working all around the world \cite{nist}. Our performances of Sec. \ref{accuracy} look poor in comparison.  As usual in big data, a huge dataset might enable us to improve our model. We are currently looking for a larger database of objects to work with. For instance, regarding the FAR we obtained, the presence of various objects which look similar   -- see, for instance, Fig. \ref{dolls} --  among the 1,000 within ALOI, tends to increase this rate in our experiments. A user does have a much larger choice at his disposal. For instance, a museum collection often counts more than a million objects while offering a long term storage for them.

\appendices
\section{Selected Model}\label{selection}
\begin{figure}[ht]
    \centering
    \includegraphics[width=0.85\columnwidth]{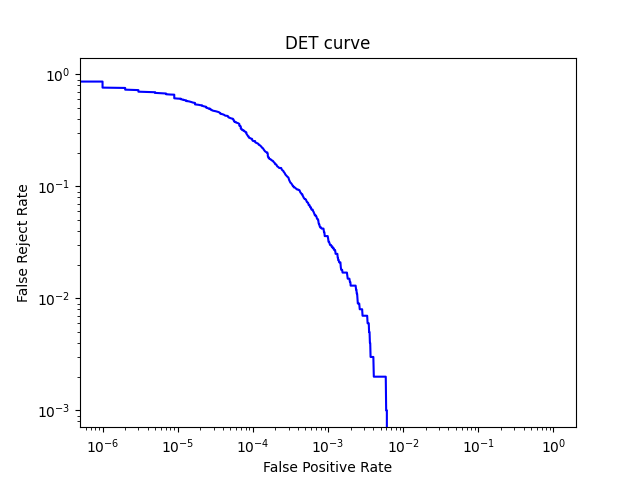}
    \caption{Rotation of 15 degrees on EfficientNet.}
    \label{fig:efficientnet_15}
\end{figure}

To turn images into binary vectors, we use an NN. We opt for the VGG architecture after various trials. For instance, we tried the pretrained EfficientNet architecture, which has a higher accuracy on ImageNet (see \url{https://paperswithcode.com/sota/image-classification-on-imagenet}). However, it yields a higher FRR for a given FAR on ALOI, as can be seen in Fig.~\ref{fig:efficientnet_15}.

\begin{figure}[ht]
    \centering
    \includegraphics[width=0.85\columnwidth]{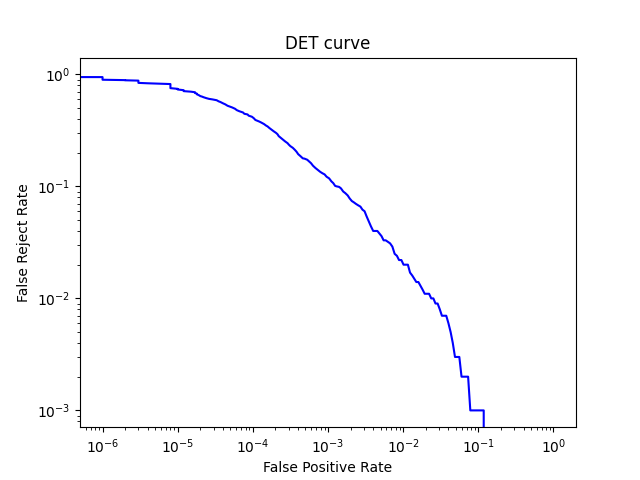}
    \caption{Rotation of 15 degrees with PCA on VGG-16.}
    \label{fig:pca_15}
\end{figure}

Similarly, in order to reduce the number of bits from 4,096 to 512, we select LSH even though other methods, such as Principal Component Analysis (PCA) exist. However, PCA depends on the training data, when LSH is independent of it. If the PCA is determined on Imagenette data rather than ALOI images, the resulting accuracy is lower than LSH, as can be seen in Fig.~\ref{fig:pca_15}. 

\section{Probability of Randomly Selecting a Template in the Hamming Ball}\label{annexe}
\cite{DBLP:conf/tcc/GalbraithZ19} determines an upper bound for the probability of randomly selecting an element $y\in\{0,1\}^n$ in ${B}_x(r)$, the Hamming ball of radius $r$ and center $x$,
taking also into account a security parameter $\lambda$.

In our case, for $n=512$, $r=140$ and $\lambda=87$, we have:  $Pr_{y\in \{0, 1\}^{512}}[y \in \mathcal{B}_x(140)] \leq \frac{1}{2^{87}}$.

Exploiting the inherent correlation between coordinates of templates, we are now going to estimate an  upper bound for $$Pr_{y\in \{0, 1\}^{n}  \cap \text{~Templates subspace}}[y \in \mathcal{B}_x(r)]$$ corresponding to when an adversary restricts himself to search among the templates subspace. This is the case when he has access to the model $M$.

Moreover in our experiments, we restrict ourselves to image inputs and we  compute the templates of the 13,394 images coming from the Imagenette dataset  (\url{https://github.com/fastai/imagenette})  searching for a false acceptance with one of the 1,000 templates of the ALOI dataset (without any rotation). 
We obtain that 4 among these 1,000 ALOI templates get a false acceptance -- a distance less than 140  -- with respectively 2, 12, 1 and 2 templates coming from the other dataset. This leads to a ratio of $1.27 \times  10^{-6}$ of the comparisons. We lose two orders of magnitude from the FAR of Sec. \ref{Implemen} due to the fact that we are here looking at images that can be quite quite different than the objects from ALOI.

\FloatBarrier
\section*{Acknowledgements.}
This work was partly supported by the iMARS project (G.A. no 883356), funded by the European Union’s Horizon 2020 research and innovation program.

\bibliographystyle{plainurl}


\bibliography{biblio_vc2}

\end{document}